**Title**

- Automatic bioinformatic software named entity recognition from literature

**Authors**

Hao Xuan[1], Rithvij Pasupuleti[1], Ben Liu[1], Haishuo Sun[1], Jun Zhang[2,3], Zijun Yao[1,*] and Cuncong Zhong[1,4,5,*]

**Affiliations**

[1]Department of Electrical Engineering and Computer Science, University of Kansas, Lawrence, KS 66045, USA
[2]OSF Healthcare Cancer Institute, Peoria, IL 61603, USA
[3]University of Illinois College of Medicine, Peoria, IL 61605 USA
[4]Bioengineering Program, School of Engineering, University of Kansas, Lawrence, KS 66045, USA
[5]Center for Computational Biology, University of Kansas, Lawrence, KS 66045, USA
[*]Correspondence: cczhong@ku.edu (C. Z.)

## Abstract

Bioinformatics software and databases are essential components of modern life science research, yet their mentions in the scientific literature are often inconsistent and difficult to systematically identify at scale. The lack of a comprehensive and up-to-date catalog of bioinformatics resources hinders efforts toward automated biomedical knowledge extraction and streamlined data analysis. Here we present SNAIL, a hybrid named entity recognition framework designed to automatically identify bioinformatics software and database (SW/DB) names from biomedical texts. SNAIL integrates complementary lexical and semantic modeling strategies. The lexical component captures orthographic patterns and contextual cues characteristic of SW/DB names, while the semantic component leverages contextual embeddings generated by transformer-based language models such as SciBERT, combined with an explicit token-masking strategy to enhance entity-focused representations. A large training corpus was constructed automatically through a hybrid pipeline that integrates citation-hinted extraction with large language model–assisted distillation. Evaluation on two independent benchmark datasets and real-world research articles demonstrates that SNAIL substantially outperforms existing approaches, including domain-specific methods such as bioNerDS2 and general-purpose large language models such as ChatGPT, Gemini, Grok and Claude. Applying SNAIL to large-scale literature analysis further reveals distinct journal-level preferences across bioinformatics subfields. These results demonstrate that SNAIL provides an accurate and scalable solution for identifying bioinformatics resources in scientific texts and enables systematic meta-analysis of tool usage and research trends.

## Introduction

Recent advances in biology and medicine have been propelled by innovations in instrumentation that enable low-cost, high-throughput, and largely unbiased acquisition of diverse biological data, including three-dimensional molecular structures (1-4), imaging datasets (5, 6), chemical compound profiles (7, 8), and genomic sequences (9, 10). The scale and complexity of these data render manual analysis infeasible, necessitating standardized computational workflows that integrate multiple specialized software tools and databases (SW/DB) (11-13). Such bioinformatic pipelines are typically assembled empirically by domain experts and are often shared through public repositories (14-16), reducing redundant effort while improving transparency and reproducibility. Despite their success, two fundamental limitations persist. First, the selection of SW/DB components is frequently guided by convention or visibility rather than systematic performance evaluation, leading to the underuse of high-quality but less prominent resources. Second, many pipelines lag behind rapid methodological advances, failing to incorporate newly developed SW/DB and consequently operating below optimal performance. Addressing these limitations requires systematic construction and continuous maintenance of a comprehensive, up-to-date catalog of bioinformatic SW/DB.

Although automatic identification of bioinformatic SW/DB from the literature might resemble a standard named entity recognition (NER) task in natural language processing (NLP), it represents a fundamentally different and more challenging problem (17-19). Conventional biomedical NER focuses on entities such as genes (20), species (21), diseases (22), drugs (23), or chemical compounds (24), which evolve slowly and are typically anchored to well-curated, stable dictionaries. By contrast, bioinformatic SW/DB are introduced at a rapid pace, making reliance on predefined vocabularies inherently insufficient and necessitating robust recognition of previously unseen entities (25-27). This challenge is further compounded by the fact that many SW/DB names are deliberately coined as common words or acronyms with unrelated everyday meanings (for example, blast (28, 29), grasp (30), or era (31), rendering surface-form features and dictionary matches unreliable for disambiguation. As a result, accurate identification requires explicit modeling of the local semantic context in which a candidate name appears, rather than dependence on spelling conventions or prior registration. Historically, these difficulties have confined SW/DB identification largely to manual curation efforts, such as Bio.tools (32) and OMICtools (33). Early automated methods such as OReFiL (34) and BIRI (35) only exploited opportunistic cues such as regular expressions or predefined word lists, failing to capture a substantial fraction of valid entities. bioNerDS (18, 36) marked a step toward automation by adopting a machine-learning framework based on capitalization patterns, syntactic features, and dictionary hits; however, its neglect of contextual semantics and reliance on a small, annotated corpus (~60 articles) limited its generalizability. More recent large language models (LLMs), including ChatGPT (37), Gemini (38), Grok (39), and Claude (40) excel at open-ended text generation but are not optimized for precise, high-recall entity classification (41, 42). Together, these limitations highlight the need for a context-aware, scalable framework that transcends dictionary-based recognition and addresses SW/DB identification as a distinct and broadly relevant information extraction problem.

To address the dual challenges of limited annotated data and insufficient modeling of semantic context, we developed SNAIL (Software NAme Identification from Literature), a hybrid framework that decouples contextual representation learning from surface-pattern recognition. SNAIL integrates a BERT-based encoder to capture the local semantic and syntactic environment in which candidate entities occur, together with a lightweight XGBoost classifier (43, 44) that models stable orthographic and dictionary-derived features, including capitalization patterns and curated name lists. We refer to these components as SNAIL-semantic and SNAIL-lexical, respectively. To enhance generalizability, lexical cues, including the SW/DB keyword itself, were explicitly masked during training of SNAIL-semantic. This design compels the model to infer entity status from contextual and syntactic signals rather than memorized token forms, thereby improving recognition of previously unseen named entities. In parallel, SNAIL-lexical captures invariant surface features characteristic of SW/DB names. Importantly, its performance can be incrementally improved through expansion of the internal dictionary, without retraining the classifier. Ablation analyses demonstrate that integrating SNAIL-semantic and SNAIL-lexical yields substantially improved performance relative to either component alone, indicating complementary signal capture. SNAIL was trained on a large corpus comprising tens of thousands of sentences, substantially expanding the effective training space beyond manually curated datasets. To mitigate the bottleneck of manual annotation, we further constructed the training corpus through a fully automated pipeline that combines citation-hinted sentence identification with large language model–assisted data distillation. This strategy enables scalable dataset generation while preserving contextual diversity. Collectively, these advances transform SW/DB name recognition from a brittle, dictionary-dependent task into a scalable, context-aware inference framework that remains robust to emerging and previously uncharacterized entities.

We benchmarked SNAIL against both a traditional machine-learning approach (bioNerDS2) and state-of-the-art general-purpose LLMs, including ChatGPT, Gemini, Grok, and Claude. Across benchmark datasets, SNAIL achieved an average F1 score exceeding 80%, substantially outperforming bioNerDS2 (35%) as well as ChatGPT (66%), Gemini (49%), Grok (58%), and Claude (28%) across the evaluated LLM variants. SNAIL processes a typical-length scientific article (~6,000 words) in approximately one minute, enabling scalable analysis of the biomedical literature. We anticipate that SNAIL will support the construction and continuous updating of a comprehensive catalog of bioinformatic SW/DB, empowering pipeline developers and end users to make informed, evidence-based tool selections and thereby improving the rigor, efficiency, and reproducibility of biomedical data analysis.

## Results

### Training and testing dataset curation

To develop a NER system for bioinformatics SW/DB recognition, large and high-quality labeled training data are essential. We constructed the SNAIL training corpus through a two-stage automated pipeline in Fig. 1. In the first stage, a citation-hinted sentence identification framework was first developed to construct high-confidence training data (Fig. 1A). A curated catalogue of bioinformatics SW/DB, each mapped to its canonical publication, was used as a seed set. Articles were scanned for exact keyword matches co-occurring with in-text citation markers, and candidate mentions were validated by cross-referencing the cited reference against the catalogue. Only citation-consistent matches were retained as positive instances, thereby minimizing false positives arising from acronym ambiguity or lexical overlap. This strategy was applied to 1,000 articles selected from PubMed (excluding the 60 articles annotated by bioNerDS2), yielding 68,864 sentences containing at least one software name. In the second stage, an LLM-assisted distillation strategy (45) was employed to further expand the training corpus (Fig. 1B). An LLM (ChatGPT) was prompted to generate labeled positive and negative examples. This automated augmentation introduces contextual diversity beyond citation-anchored instances. Through an LLM-guided distillation pipeline, we generated a total of 50,000 annotated sentences, including 40,000 positive sentences and 10,000 adversarial negative sentences. The citation-hinted and LLM-generated datasets were subsequently merged to construct the final SNAIL training corpus, which contains 134,681 positive tokens and 2,441,518 negative tokens.

The model was evaluated on two independent manually annotated datasets, neither of which shared articles with the training corpus. The first dataset (DS1) consists of 148,131 sentences annotated in this study, comprising 131,209 positive tokens and 2,967,666 negative tokens. The second dataset (DS2) includes 60 articles (8,196 sentences) previously annotated by bioNerDS2, containing 1,325 positive tokens and 195,390 negative tokens. The training dataset, DS1, and DS2 are publicly available through the URLs provided in the Data and materials availability (see Supplementary Data 1).

**Overview of SNAIL architecture**

SNAIL is a hybrid framework that consists of two complementary components: SNAIL-lexical and SNAIL-semantic (Fig. 2). The first component, SNAIL-lexical, captures invariant surface-level features of SW/DB names, including capitalization patterns, abbreviation structures, and curated dictionary signals, using an XGBoost classifier. The second component, SNAIL-semantic, models the local contextual environment of each candidate token using a transformer-based language encoder, enabling recognition of previously unseen entities based on surrounding syntactic and semantic cues. The outputs of the two components are subsequently combined through a late-fusion strategy to produce the final token-level predictions. Text preprocessing and tokenization are performed using NLTK (46) and spaCy (46). A detailed description of the SNAIL framework and implementation is provided in Materials and Methods.

**SNAIL model selection**

Because the SNAIL training corpus consists of both citation-derived and LLM-generated examples, model selection was performed under three training configurations: citation-hinted sentences alone (CIT), LLM-generated sentences alone (LLM), and the merged dataset containing both sources (Both). This design enabled us to assess the impact of training data composition on both lexical and semantic model performance.

To identify the most suitable classifier for SNAIL-lexical, we evaluated multiple supervised learning algorithms, including support vector machines (SVM), logistic regression, random forest, multilayer perceptron (MLP), and XGBoost (44). Across all classifiers, models trained on the merged dataset consistently outperformed those trained on either source alone (Fig. 3A and Fig. S1A), providing a substantial F1-score increase of up to 9.7% over the citation-hinted subset and up to 6.4% over the LLM-generated synthetic dataset. Among the evaluated classifiers, XGBoost achieved the best overall performance on both DS1 and DS2 using merged datasets, yielding peak F1-scores of 84.6% (DS1) and 82.5% (DS2), and was therefore adopted as the SNAIL-lexical classifier.

To explore the most suitable semantic embedding model for SNAIL-semantic, we evaluated four embedding approaches: TF-IDF (47), FastText (48), bioBERT (49), and SciBERT (50).

Consistent with the lexical classification results, the merged dataset again yielded the best performance across all embedding strategies in Fig. 3B and Fig. S1B, providing an F1-score boost of up to 7.0% over the citation-hinted subset and up to 5.7% over the LLM-generated synthetic data. Among the evaluated embeddings, transformer-based models achieved the highest overall performance on both DS1 and DS2 using merged datasets, led by SciBERT with peak F1-scores of 87.1% (DS1) and 86.4% (DS2), followed by BioBERT at 75.8% (DS1) and 75.0% (DS2), whereas the next-best approach, FastText, remained below 70%. These results demonstrate that pre-trained contextual language models capture richer semantic representations for SW/DB recognition than traditional embedding methods.

Because transformer-based models may rely on the surface form of software names, we investigated whether explicitly masking the target token could encourage the model to learn contextual patterns from the surrounding text. Since SciBERT and BioBERT produced the most accurate semantic embeddings across both DS1 and DS2, the masking strategy was evaluated using these two encoders. The results indicate that explicit masking of the target token consistently improves performance for both models, yielding 6-10% higher F1-scores compared with models trained without masking (Fig. 3C and Fig. S1C). Accordingly, the final SNAIL-semantic model employs a SciBERT encoder with the explicit masking strategy.

SNAIL-lexical and SNAIL-semantic capture complementary sources of information: lexical surface patterns and contextual semantics, respectively. To quantify the benefit of integrating both components, we performed an ablation study comparing the full SNAIL model against each component in isolation. As shown in Fig. 3D and Fig. S1D, the integrated model consistently outperformed its ablation components, achieving peak F1-scores of 94.8% on DS1 and 94.4% on DS2, which justifies the integration of both lexical and semantic features.

**Performance benchmark**

We benchmarked the final SNAIL model against bioNerDS2 on both DS1 and DS2. To our knowledge, bioNerDS2 is the only previously reported method specifically designed to identify bioinformatics SW/DB named entities. The detailed outputs of both SNAIL and bioNerDS2 are available via the URLs provided in the Data and materials availability. As summarized in Fig. 4A and Fig. S2, SNAIL substantially improves performance, increasing recall, precision and F1-score from approximately 62% for bioNerDS2 to over 90%.

Finally, we compared the performance of SNAIL with several contemporary large language models (LLMs), including earlier-generation and more recent variants, such as Claude, Gemini, Grok, and GPT (see Table S1). Owing to computational constraints, this comparison was conducted on two manually annotated articles (PMC6082860 (13) and PMC12857227 (51)), rather than on the larger benchmark datasets DS1 and DS2. Each sentence from the selected articles was provided to the LLMs, which were prompted with the instruction: "Identify all bioinformatics software, methods and/or database named entities in the given sentence." The manual annotations of the two articles, together with the outputs of SNAIL and the evaluated LLMs, are available via the URLs provided in the Data and materials availability (see Supplementary Data 1).

Performance was quantified with all reported values corresponding to the average across the two annotated articles (Fig. 4B). Across these evaluations, SNAIL consistently achieved the highest performance, with an average F1-score of approximately 82-83%, substantially outperforming all evaluated LLMs. Among the LLMs, performance improved with newer model generations, with GPT-5.4-mini achieving the best results (average F1 ≈ 69-71%), followed by earlier models such as GPT-4.1-mini (average F1 ≈ 63%) and Grok variants. Notably, although LLMs generally exhibited high recall (often exceeding 80-90%), their precision remained substantially lower, leading to reduced overall F1-scores. In contrast, SNAIL maintains a more balanced trade-off between precision and recall, resulting in superior overall accuracy.

The close agreement between the two articles indicates that these averaged results reflect stable model behavior rather than article-specific variation. Together, these findings highlight a fundamental distinction between task-specific models and general-purpose LLMs: while LLMs are effective at broadly identifying candidate entities, specialized architectures such as SNAIL are better suited for precise, high-confidence named entity recognition in domain-specific contexts.

**Applications on large-scale literature mining**

Finally, we applied SNAIL to a large literature corpus to assess its performance in real-world settings. A total of 2,000 scientific articles were randomly sampled from ten leading bioinformatics journals, including Bioinformatics, BMC Bioinformatics, Nature Biotechnology, Nature Methods, PLOS Computational Biology, Genome Biology, Genome Research, Nucleic Acids Research, Briefings in Bioinformatics, and IEEE/ACM Transactions on Computational Biology and Bioinformatics (TCBB), with approximately 200 articles drawn from each journal. The PubMed IDs of the 2,000 selected articles are available via the URLs provided in the Data and materials availability (see Supplementary Data 1).

From this corpus, we identified the 22 most frequently mentioned SW/DB named entities. Manual inspection confirmed that all correspond to genuine bioinformatics resources. Nine entities were subsequently excluded because they lack a unique or identifiable citation source. These include “R”, “ROC”, “Cluster”, “S4”, “Sigma”, “GenBank” (52), “Prism” (53), “Ensembl” (54) and “MIA” (55).

For the remaining 13 SW/DB NEs, a strong positive correlation was observed between their mention frequencies in the literature and their citation counts (Pearson correlation coefficient = 0.8; Spearman rank correlation coefficient = 0.8; p-value <= 0.001; Fig. 5A). Together, these results indicate that SNAIL can reliably identify bioinformatics SW/DB entities from large-scale scientific literature.

We next investigated whether different journals exhibit distinct preferences for specific bioinformatics subfields. To this end, we performed hierarchical clustering based on the mention frequencies of the 13 most frequently referenced SW/DB NEs across the ten selected bioinformatics journals. Several meaningful clusters emerged from this analysis (Fig. 5B). Notably, KEGG (56) and Gene Ontology (GO) (57) clustered together, reflecting their shared roles in gene function annotation and pathway analysis. Similarly, Protein Data Bank (PDB) and DrugBank (58) formed another cluster, consistent with their frequent joint use in studies of protein structure and protein–small molecule interactions, particularly in drug discovery and virtual screening (1, 59). Overall, the clustering patterns suggest that research on gene function organization and protein three-dimensional structure represent two of the most prominent themes across the selected journals. Journal-level preferences are also evident. For instance, IEEE/ACM TCBB and Briefings in Bioinformatics cluster together, largely driven by their relatively high mention frequencies of Protein Data Bank, suggesting a shared emphasis on protein structural analysis. In contrast, PLOS Computational Biology, Nucleic Acids Research, and Nature Methods show substantially higher mention frequencies of GO relative to PDB, indicating a stronger focus on gene function annotation and functional genomics.

**Discussion**

The rapid expansion of bioinformatics tools and databases has made computational resources indispensable to modern life science research. However, there exists no comprehensive and up-to-date catalog for such bioinformatic resources. To facilitate the construction of such a catalog, we developed SNAIL, a hybrid named entity recognition framework designed to automatically identify bioinformatics SW/DB mentions from biomedical texts. By integrating lexical pattern recognition with contextual semantic modeling, SNAIL achieves robust recognition performance across diverse writing styles and publication venues.

The design of SNAIL highlights the complementary strengths of lexical and semantic approaches for domain-specific entity recognition. The lexical component effectively captures orthographic patterns that are common in SW/DB names, such as capitalization patterns, abbreviations, and token structures. Meanwhile, the semantic component leverages contextual embeddings generated by transformer-based language models such as SciBERT to incorporate surrounding textual context. In addition, the explicit token-masking strategy further improves performance by encouraging the model to focus on contextual cues surrounding the candidate entity.

Another key contribution to this work is the automated construction of a large training corpus. Rather than relying solely on manually curated annotations, which are expensive and difficult to scale, we combined citation-hinted extraction with large language model–assisted distillation to generate high-quality training data. This hybrid strategy enables the efficient generation of large, labeled datasets while maintaining sufficient annotation quality for training accurate models. The results demonstrate that this approach can support the development of effective domain-specific entity recognition systems. The benefit of this hybrid training strategy was not limited to dataset scale. A notable finding from our component-level analysis was that merging CIT and LLM data improved performance across both SNAIL-lexical and SNAIL-semantic—two components with fundamentally different learning paradigms. The consistency of this pattern across model architectures suggests that real and synthetic examples capture complementary dimensions of domain-specific contexts. Citation-grounded sentences anchor the model in realistic usage patterns with reduced annotation noise, while LLM-generated sentences introduce broader linguistic diversity and adversarial contexts that help models generalize across heterogeneous writing styles. These findings align with recent observations that synthetic augmentation is most effective when it expands coverage into regions of the input space underrepresented in real data (60, 61). For future practitioners, this implies that even modest amounts of high-quality synthetic data—when carefully curated—can meaningfully enhance model performance beyond what either data source alone can achieve.

Our benchmarking experiments show that SNAIL consistently outperforms existing approaches, including a traditional random forest classifier bioNerDS2 and several large language models such as ChatGPT, Gemini, Grok, and Claude, when applied to the specific task of identifying bioinformatics SW/DB names. This result highlights an important distinction between general-purpose language models and specialized information extraction systems. While large language models possess broad linguistic knowledge, task-specific architectures trained on domain-tailored datasets can still provide superior accuracy for specialized biomedical text mining tasks.

Beyond entity recognition, the application of SNAIL to large-scale literature analysis demonstrates its value for meta-research in bioinformatics. By automatically identifying SW/DB mentions across thousands of publications, SNAIL enables systematic analysis of tool usage patterns across journals and subfields. For example, our analysis revealed distinct journal-level preferences for different categories of bioinformatics resources, including gene function annotation resources such as GO and KEGG, as well as structural biology resources such as PDB. Such analyses provide a quantitative perspective on how computational resources are adopted and disseminated within the bioinformatics community.

Future work could extend SNAIL in several directions. Incorporating additional contextual signals, such as citation patterns, SW/DB usage statements, version number, document-level mention frequency, may further improve entity recognition accuracy. The framework could also be expanded to extract richer information about bioinformatics resources, including their functional categories, dependencies, and usage relationships. Ultimately, integrating such information into a continuously updated knowledge base could facilitate the construction of a comprehensive catalog of bioinformatics tools and databases.

In summary, SNAIL provides an accurate and scalable framework for automatically identifying bioinformatics SW/DB in biomedical literature. By combining lexical pattern recognition with contextual semantic modeling and an automated training corpus construction strategy, SNAIL enables reliable large-scale extraction of computational resources from scientific texts. The ability to systematically track SW/DB usage across publications opens new opportunities for quantitative meta-research, including the analysis of tool adoption, methodological trends, and the evolution of bioinformatics subfields. More broadly, SNAIL illustrates how the hybrid machine learning architectures and automated data distillation pipelines can support the development of domain-specific knowledge extraction systems. As biomedical literature continues to grow rapidly, such approaches will become increasingly important for enabling autonomous knowledge mining and for maintaining comprehensive, up-to-date catalogs of computational resources used in life science research.

## Materials and Methods

### SNAIL-lexical

SNAIL-lexical incorporates 13 hand-engineered features largely adapted from bioNerDS2. These features fall into three categories: lexical attributes, dictionary-matching indicators and syntactic context features (Table 1). The lexical attributes were directly computed from the keyword. Syntactic features include detection of Hearst patterns, implemented using spaCy, and identification of enumeration structures, defined as noun phrases separated by commas and coordinating conjunctions. Dictionary-based features were derived from multiple curated resources. The “Good headword,” “Weak headword,” and “Blacklist headword” dictionaries were obtained from bioNerDS2. The “known bioinformatics

SW/DB” dictionary was also adapted from bioNerDS2 and further expanded with approximately 200 manually curated entries. The Bioconductor resource list was retrieved from its official website, and biological acronyms were obtained from the bioarco database (62). General-language controls were included using the SCOWL (63) English word list and the AcronymFinder database (64) to flag common English words and acronyms.

To identify the best model for SNAIL-lexical, we evaluated five supervised classifiers, including support vector machine (SVM), logistic regression, random forest, multilayer perceptron (MLP), and extreme gradient boosting (XGBoost). Class labels were encoded using LabelEncoder. Model selection was based on the F1 score.

For SVM, class imbalance was addressed using balanced class weights. Hyperparameters were optimized using GridSearchCV with the regularization parameter $C \in \{0.1, 1, 10\}$ and kernel coefficient $\gamma \in \{scale,\ 0.01, 0.1, 1\}$ with a radial basis function (RBF) kernel. The optimal configuration used $C = 10$ and $\gamma = 0.01$.

For logistic regression, we implemented a pipeline combining feature scaling and classification. Features were standardized using StandardScaler ($with_mean = False$) followed by L2-regularized logistic regression with the liblinear solver and balanced class weights. Hyperparameters were tuned over $C \in \{0.001, 0.01, 0.1, 1, 10, 100\}$. The best-performing model used $C = 0.1$.

For random forest, hyperparameters were optimized using grid search over the number of trees $(100, 200, 300)$, maximum tree depth $(None, 10, 20)$, and feature sampling strategy (sqrt). The best configuration consisted of 100 trees without depth restriction and sqrt feature sampling.

For MLP, we implemented an imblearn pipeline to accommodate the training data imbalance issue. The pipeline consists of StandardScaler ( $with_mean = False$ ), RandomOverSampler, and an MLPClassifier with two hidden layers $(32, 32)$, ReLU activation, the Adam optimizer, adaptive learning rate scheduling, early stopping, and a maximum of 300 training iterations. Hyperparameters were optimized using grid search over the initial learning rate $(0.0001, 0.001, 0.01)$ and L2 regularization parameter $\alpha \in \{0.0001, 0.01, 0.1\}$. The optimal model used $\alpha = 0.1$ and an initial learning rate of 0.01.

Finally, we implemented an XGBoost model incorporating both L1 ($\alpha = 1.0$) and L2 ($\lambda = 2.0$) regularization and was optimized over the learning rate $(0.01, 0.1, 0.5)$, maximum tree depth $(2, 3, 4)$, and number of boosting rounds $(25, 50, 75)$. The best baseline configuration used a learning rate of 0.5, maximum depth of 4, and 25 boosting rounds. Further fine-tuning explored additional parameters including positive class weight scaling, learning rate, tree depth, and number of estimators, evaluating 625 parameter combinations with early stopping. The final optimized model used a class weight scaling factor of 0.6, learning rate 0.45, maximum depth 4, and 15 boosting rounds.

**SNAIL-semantic**

SNAIL-semantic employs an architecture in which contextual embeddings are first generated from the input text fragment, followed by a classification layer. To encourage reliance on contextual and syntactic signals rather than memorized surface forms, the target keyword was explicitly masked during embedding generation. Embeddings were computed over a fixed-width context window comprising five upstream and five downstream tokens surrounding the masked position. The resulting representation was then classified using a fully connected layer with 2,048 hidden units.

We evaluated four embedding approaches, including TF-IDF, FastText, BioBERT, and SciBERT. To explore the most effective semantic representation for software and database mention recognition, we evaluated four widely used text embedding approaches, including two transformer-based biomedical language models (SciBERT and BioBERT) and two traditional embedding methods (FastText and TF–IDF).

For fair comparison, all embeddings were coupled with the same neural classification architecture consisting of a fully connected network with a single hidden layer of 2,048 units, followed by a GELU activation function and a dropout layer ($rate = 0.1$). The final output layer produced the probability that a given token corresponds to a software or database mention. All models were trained using binary cross-entropy loss and optimized with the Adam optimizer ($\varepsilon = 1 \times 10^{-8}$, $weight\ decay = 0.01$). Training was performed using mixed-precision computation (FP16) with the native automatic mixed precision (AMP) backend. Gradient clipping with a maximum norm of 1.0 was applied to stabilize optimization. A polynomial decay learning-rate scheduler gradually decreased the learning rate to a minimum value of $3 \times 10^{-8}$ during training. Hyperparameters were optimized using grid search.

For contextual biomedical language modeling, we implemented transformer-based encoders using SciBERT (scivocab_uncased) and BioBERT v1.1 pretrained on PubMed abstracts. Both models follow the BERT-base architecture with 12 transformer layers, 12 self-attention heads, and a hidden size of 768, producing contextualized token embeddings for each input sentence. Input sequences were truncated or padded to a maximum length of 200 tokens. The contextual embeddings were passed to the neural classifier described above. An additional dropout layer ($\mathrm{rate} = 0.3$) was applied to the masked-token prediction layer to improve generalization. The best-performing configuration used an initial learning rate of $3 \times 10^{-8}$ for both SciBERT and BioBERT.

We implemented a model using TF–IDF features as input representations. TF–IDF vectors were generated using the TfidfTransformer implementation from scikit-learn (v0.24.2) (*65*), trained on the full training corpus to capture token importance across the dataset. The resulting sparse feature vectors were used as input to the neural classifier. The optimal initial learning rate ($1 \times 10^{-6}$) was selected through grid search.

We also evaluated FastText embeddings as an alternative semantic representation. FastText was selected instead of Word2Vec (66) because its subword modeling enables vector representations for rare or out-of-vocabulary tokens, which are common in bioinformatics software names. Pretrained FastText vectors trained on the first 1 billion bytes of English Wikipedia were used to obtain word-level embeddings for each sentence. The resulting sentence representations were then passed to the neural classifier. The optimal initial learning rate (0.5) was determined through grid search.

**Model fusion**

SNAIL-lexical and SNAIL-semantic are subsequently integrated to form the final SNAIL model, enabling joint consideration of lexical and semantic signals. SNAIL-lexical implements an XGBoost classifier that produces $n$ tree-level probability estimates $p_1, p_2, \ldots, p_n$ for a target token, where $n$ denotes the number of decision trees in the ensemble. These probabilities are first ranked and then aggregated using exponentially decayed positional weights:

$$w_i = \frac{\exp(-\lambda r_i)}{\sum_{j=1}^{n} \exp(-\lambda r_j)}$$

where $r_i$ denotes the rank of tree $i$, and $\lambda$ controls the decay rate. The weighted mean probability and its corresponding variance are computed as:

$$\mu = \sum_{i=1}^{n} w_i \, p_i$$

and

$$\sigma^2 = \sum_{i=1}^{T} w_i \, (p_i - \mu_i)^2.$$

The variance $\sigma^2$ serves as an estimate of ensemble uncertainty, allowing derivation of a confidence modulation factor that down-weights inconsistent rule activations.

Because XGBoost outputs probabilities in the interval [0,1], whereas SNAIL-semantic (SciBERT) produces real-valued logits, the aggregated lexical probability $\mu$ is mapped to logit space via a log-odds transformation to ensure numerical compatibility between the two prediction sources. Final predictions are obtained through additive fusion in logit space:

$$l^{fused} = \beta l^{semantic} + (1-\beta)\exp(-\gamma\sigma^2)\log\frac{\mu}{1-\mu}$$

where $l^{\text{fused}}$ denotes the fused logit, and $l^{\text{semantic}}$ is the logit produced by SNAIL-semantic. The parameter $\beta \in [0,1]$ balances semantic and lexical contributions, whereas $\gamma > 0$ controls sensitivity to ensemble disagreement.

**Benchmark experiment and performance metrics**

Model performance was evaluated at the mention level using precision, recall, and F1-score. True positives (TP) were defined as correctly identified SW/DB named entities. False positives (FP) correspond to non–SW/DB mentions incorrectly predicted as entities, whereas false negatives (FN) denote SW/DB entities missed by the model. Performance estimates were obtained using five-fold stratified cross-validation on the training corpus, and results are reported as mean ± s.d. across folds. All experiments were implemented in PyTorch (v1.7.1) with CUDA (v11.0.221) and executed on NVIDIA A100 GPUs.

## Acknowledgments

**Funding:** This work was supported by U.S. National Science Foundation (https://ror.org/021nxhr62; DBI-1943291). All authors read and approved of the final manuscript.

**Author contributions:**
Conceptualization: HX, CZ
Methodology: HX, CZ, ZY
Software: HX
Formal analysis: HX
Investigation: HX, RP
Visualization: HX
Supervision: HX, CZ
Writing—original draft: HX, CZ
Writing—review & editing: JZ, ZY, CZ, HX

**Competing interests:** Authors HX, JZ, and CZ hold equity interests in H2Alpha Inc., a company developing technologies related to the subject of this study. This relationship had no role in the study design, data collection, analysis, interpretation, or manuscript preparation. All other authors declare that they have no competing interests.

**Data and materials availability:** All datasets and supplementary materials used in this study are publicly available on Figshare (https://doi.org/10.6084/m9.figshare.31558663). A comprehensive index of these files with interactive download links is available in Supplementary Data 1. The repository contains:

(1) The training and evaluation datasets (DS1 and DS2).
(2) Outputs of SNAIL and bioNerDS2 on the DS1 and DS2 benchmark datasets.
(3) Manual annotations of two articles used to evaluate SNAIL and large language models (LLMs).
(4) Outputs of SNAIL and the evaluated LLMs on the two manually annotated articles.
(5) PubMed identifiers (PMIDs) of the 2,000 articles used for large-scale literature mining.

**Figures and Tables**

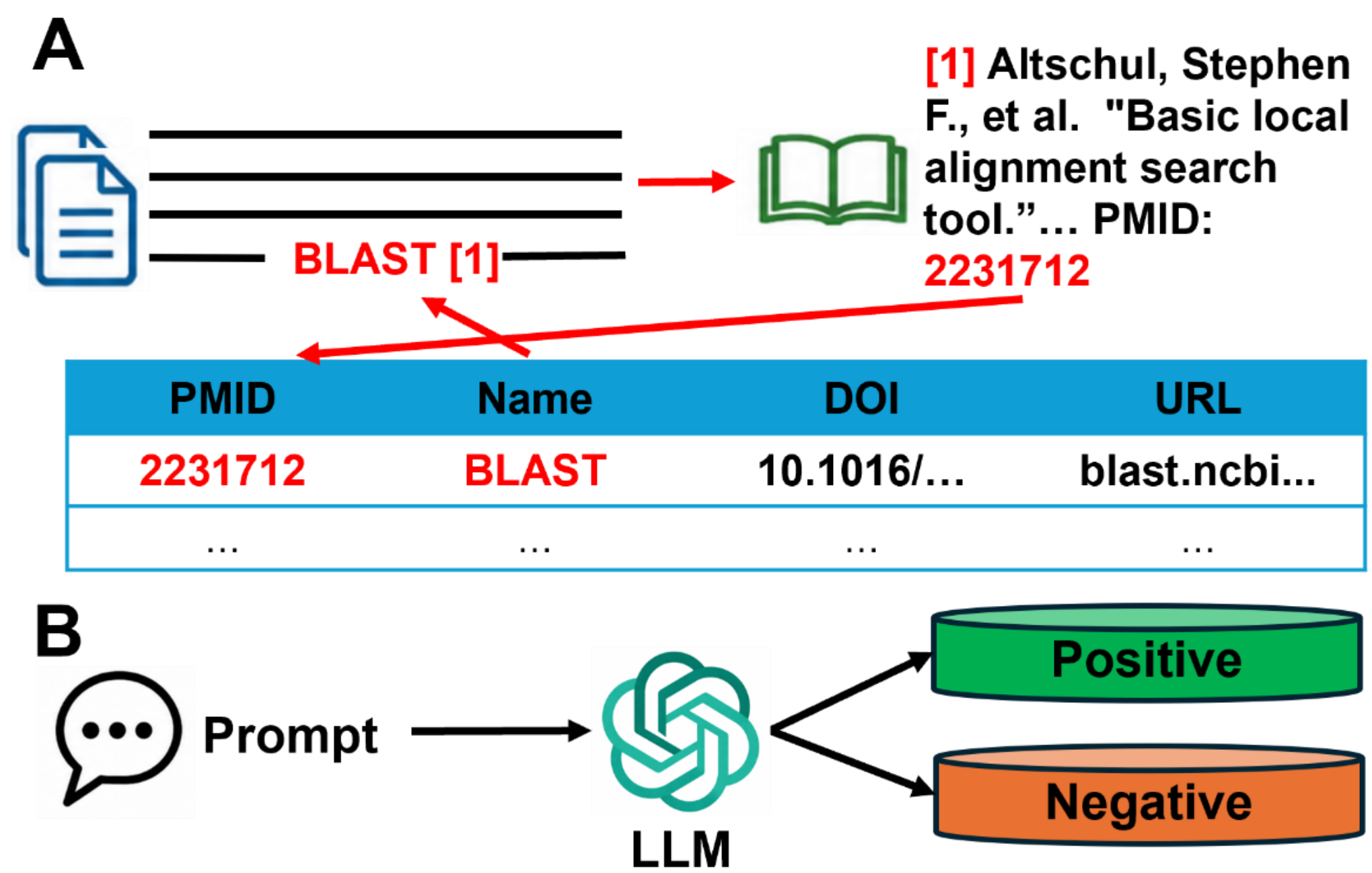

**Fig. 1. Overview of the automated training corpus generation framework.** The workflow construction integrates literature mining with generative data augmentation. (**A**) Citation-hinted extraction**.** Sentences containing software/database (SW/DB) references are isolated and validated using publication metadata. (**B**) LLM-assisted generation**.** Tailored prompts guide a large language model to distill diverse positive and negative training examples from the extracted text.

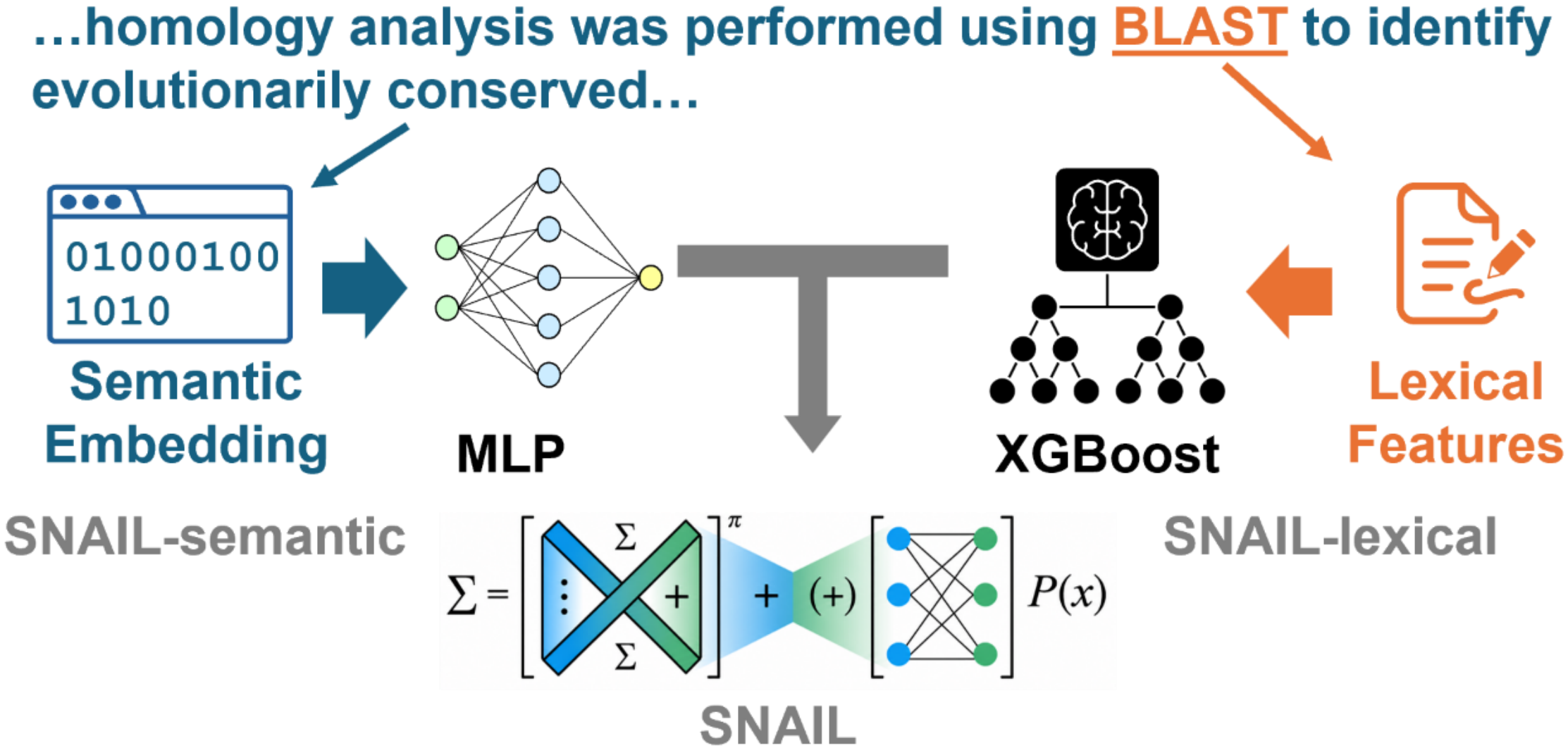

**Fig. 2. Hybrid architectural design of the SNAIL framework.** The framework combines semantic and lexical processing tracks for token-level prediction. The semantic stream (SNAIL-semantic) processes text embeddings through an MLP, while the lexical stream (SNAIL-lexical) routes engineering features through XGBoost. Output vectors from both streams are integrated via a late-fusion layer.

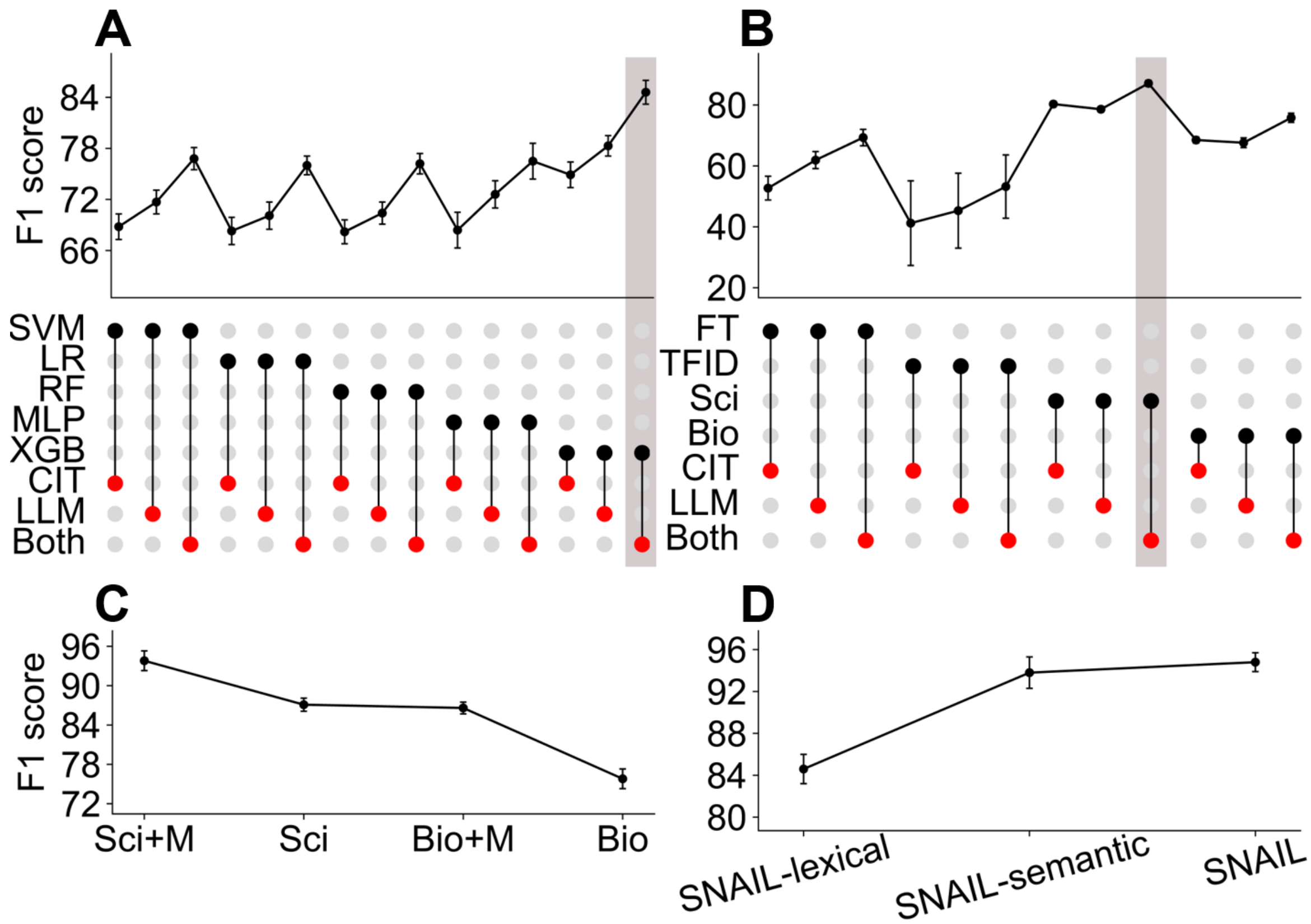

**Fig. 3. Model selection, ablation analysis, and integration of SNAIL components across datasets and training strategies on DS1.** (**A**) Comparison of lexical classifiers for SNAIL-lexical. Models include support vector machine (SVM), logistic regression (LR), random forest (RF), multilayer perceptron (MLP), and extreme gradient boosting (XGBoost, XGB). Black dots indicate the selected lexical classifier, whereas red dots indicate the training dataset configuration: citation-hinted sentences (CIT), large language model–generated sentences (LLM), or the merged dataset containing both sources (Both). The upper panel shows the corresponding F1-scores. The vertical gray shaded bar highlights the optimal configuration. (**B**) Comparison of semantic embedding strategies for SNAIL-semantic using FastText (FT), term frequency–inverse document frequency (TF-IDF, TFID), SciBERT (Sci), and BioBERT (Bio) embeddings coupled with an MLP classifier. Black dots denote the selected embedding model, and red dots denote the training dataset configuration (CIT, LLM, or Both). The upper panel shows the corresponding F1-scores. The vertical gray shaded bar highlights the optimal configuration. (**C**) Effect of explicit masking of the candidate token during semantic model training. Performance is compared for SciBERT (Sci) and BioBERT (Bio) models trained with masking (Sci+M and Bio+M) or without masking (Sci and Bio). (**D**) Performance comparison of the integrated framework components, including SNAIL-lexical, SNAIL-semantic, and the final fused SNAIL model. Error bars represent standard deviation across cross-validation folds.

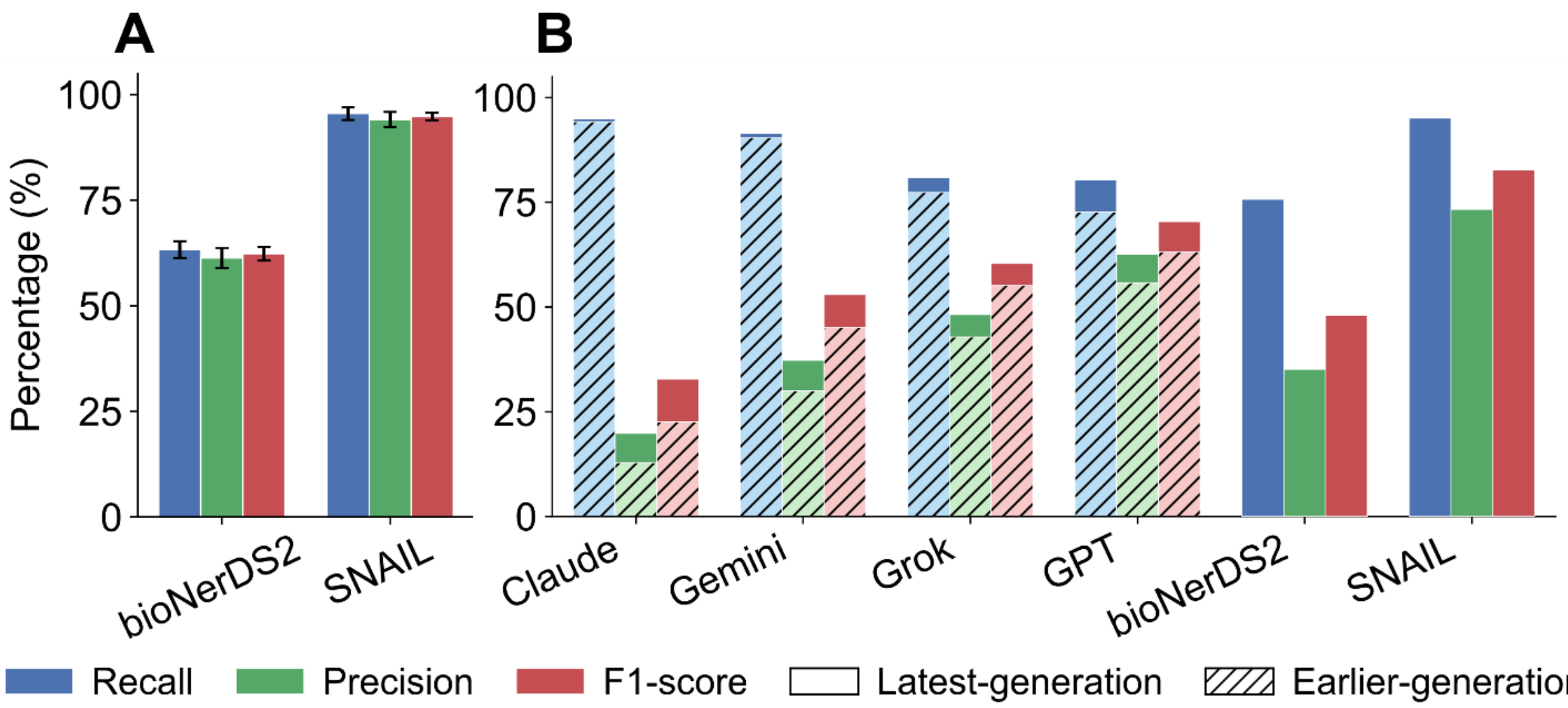

**Fig. 4. Benchmarking of SNAIL against existing methods and large language models for bioinformatics named entity recognition.** (**A**) Performance comparison on benchmark datasets. Precision, recall, and F1-score of SNAIL and bioNerDS2 evaluated on DS1. Error bars indicate standard deviation across cross-validation folds. (**B**) Comparison with large language models (LLMs) on two full-text articles (PMC6082860 and PMC12857227). Each sentence was evaluated for software/database entity recognition using SNAIL, bioNerDS2, and representative LLMs. Hatched bars denote earlier-generation models (Claude-3.5-Sonnet, Gemini-2.5-Flash, Grok-4, GPT-4.1-mini), and solid bars denote more recent models (Claude-Sonnet-4.6, Gemini-3-Flash-preview, Grok-4.20-0309-reasoning, GPT-5.4-mini).

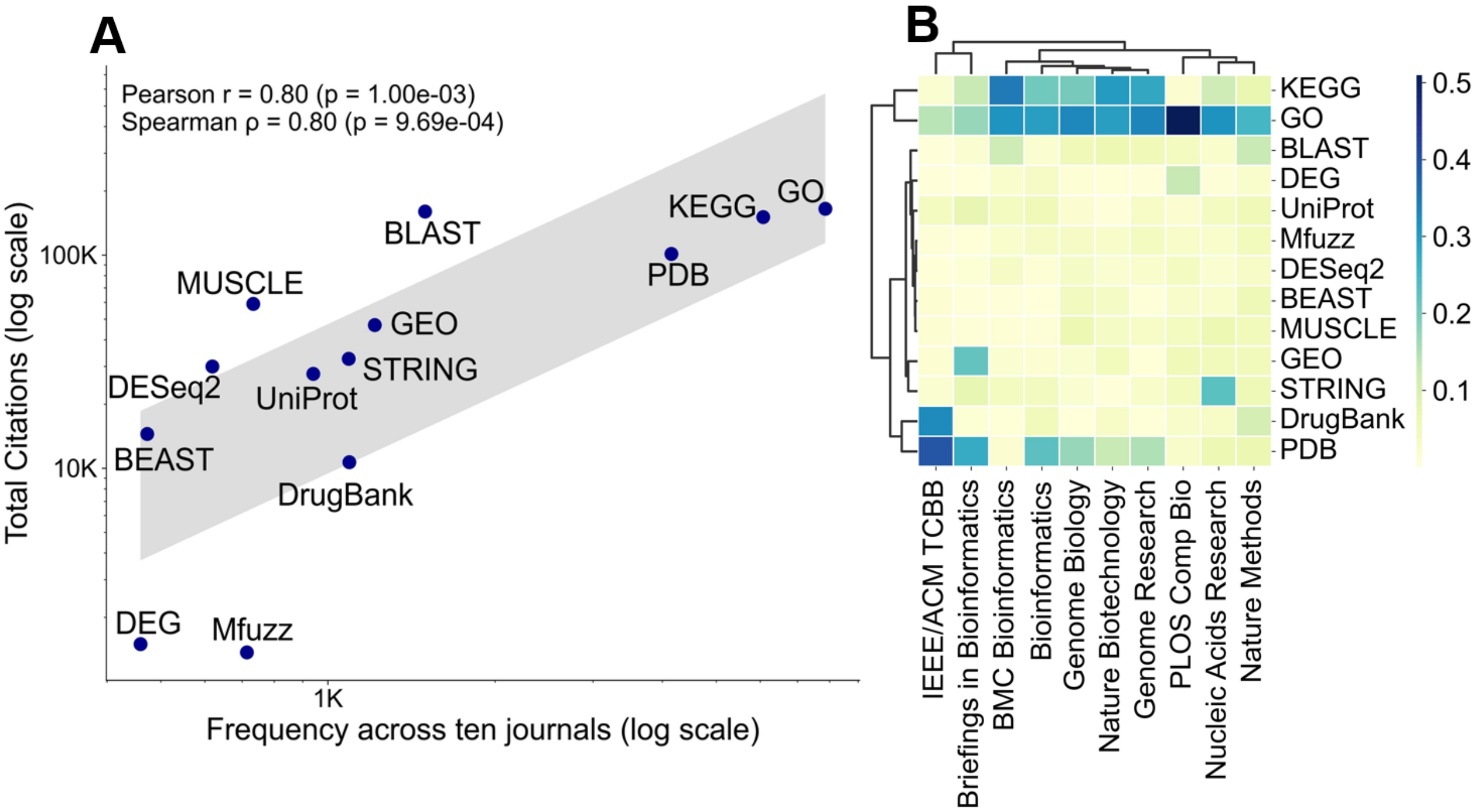

**Fig. 5. Large-scale literature mining reveals correlations between tool usage and citation impact, and journal-specific preferences in bioinformatics resources.** (**A**) Correlation between mention frequency and citation counts. Each point represents a bioinformatics software or database identified by SNAIL across 2,000 articles from ten journals. (**B**) Journal-level clustering based on tool usage profiles. Heatmap and hierarchical clustering of normalized mention frequencies for the most frequently identified software and databases across ten bioinformatics journals.

**Table 1: A summary of all features used in the SNAIL-lexical model.** Features are grouped into lexical, syntactic, and dictionary-based categories, capturing naming patterns, contextual cues, and prior knowledge of bioinformatics software and databases (SW/DB).

| Feature | Category | Example/Note |
|---|---|---|
| Upper-case | Lexical | BLAST, PDB |
| Lower-case | Lexical | blastp, nr, nt |
| Mixed-cased | Lexical | edgeR, DESeq2 |
| Hearst pattern | Syntactic | "…tools such as BLAST…" |
| Enumeration | Syntactic | "…such as BWA, Bowtie, and SOAP…" |
| Good headword | Dict. Match | database, tools |
| Weak headword | Dict. Match | platform, interface |
| Blacklist headword | Dict. Match | algorithm, method |
| Bioconductor | Dict. Match | a list of known Bioconductor packages |
| Known SW/DB | Dict. Match | a list of known bioinformatic SW/DB NEs |
| Biological acronyms | Dict. Match | a list of biochemical reagents |
| English words | Dict. Match | a list of English words |
| English acronyms | Dict. Match | a list of English acronyms |

**Supplementary Materials**
Fig. S1 to S2
Table S1

# Supplementary Materials for

## Automatic bioinformatic software named entity recognition from literature

Hao Xuan *et al.*

*Corresponding author. Email: cczhong@ku.edu

**This PDF file includes:**

Fig. S1 to S2
Table S1

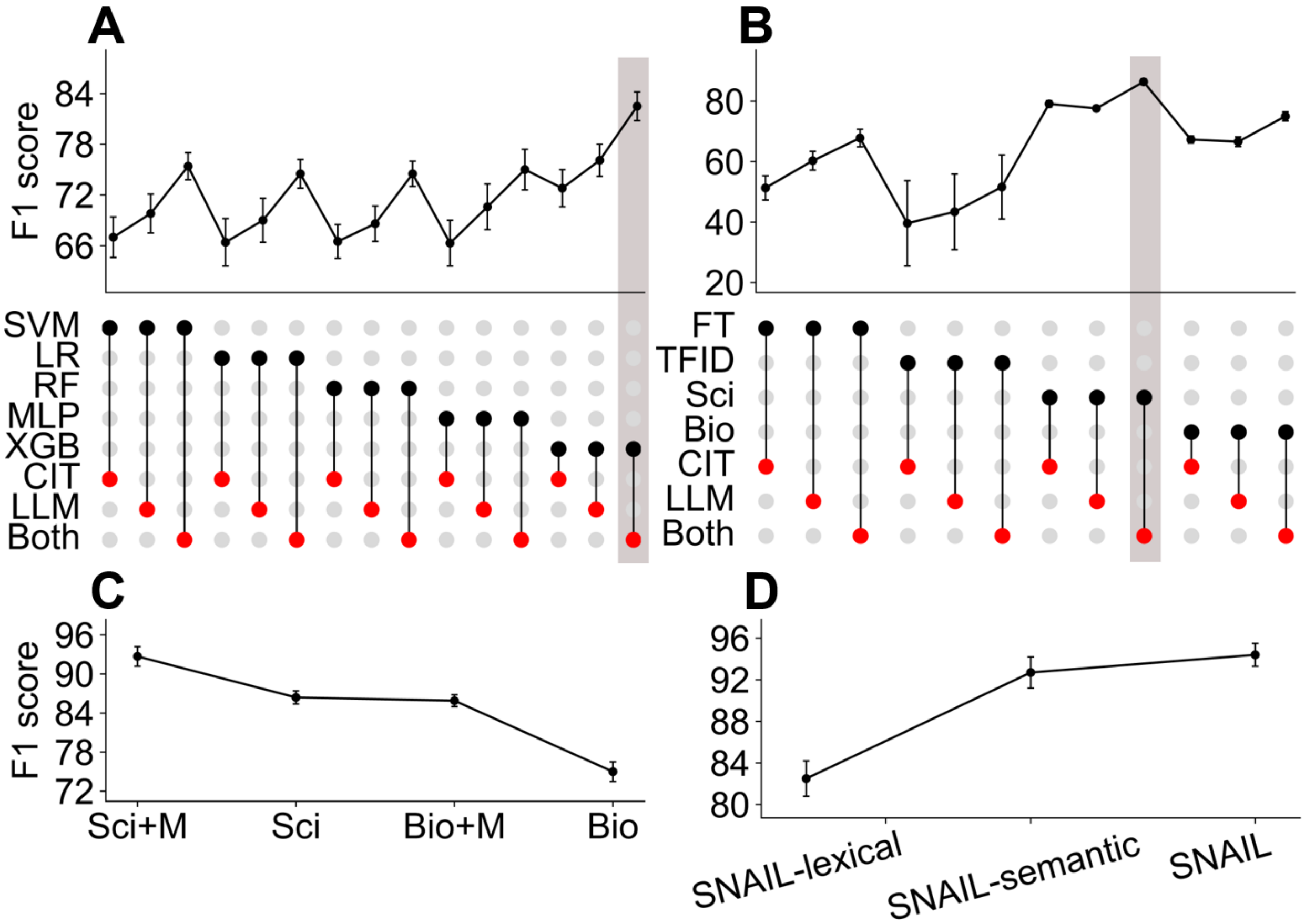

**Fig. S1: Model selection, ablation analysis, and integration of SNAIL components across datasets and training strategies on DS2.** (**A**) Comparison of lexical classifiers for SNAIL-lexical. Models include support vector machine (SVM), logistic regression (LR), random forest (RF), multilayer perceptron (MLP), and extreme gradient boosting (XGBoost, XGB). Black dots indicate the selected lexical classifier, whereas red dots indicate the training dataset configuration: citation-hinted sentences (CIT), large language model–generated sentences (LLM), or the merged dataset containing both sources (Both). The upper panel shows the corresponding F1-scores. The vertical gray shaded bar highlights the optimal configuration. (**B**) Comparison of semantic embedding strategies for SNAIL-semantic using FastText (FT), term frequency–inverse document frequency (TF-IDF, TFID), SciBERT (Sci), and BioBERT (Bio) embeddings coupled with an MLP classifier. Black dots denote the selected embedding model, and red dots denote the training dataset configuration (CIT, LLM, or Both). The upper panel shows the corresponding F1-scores. The vertical gray shaded bar highlights the optimal configuration. (**C**) Effect of explicit masking of the candidate token during semantic model training. Performance is compared for SciBERT (Sci) and BioBERT (Bio) models trained with masking (Sci+M and Bio+M) or without masking (Sci and Bio). (**D**) Performance comparison of the integrated framework components, including SNAIL-lexical, SNAIL-semantic, and the final fused SNAIL model. Error bars represent standard deviation across cross-validation folds.

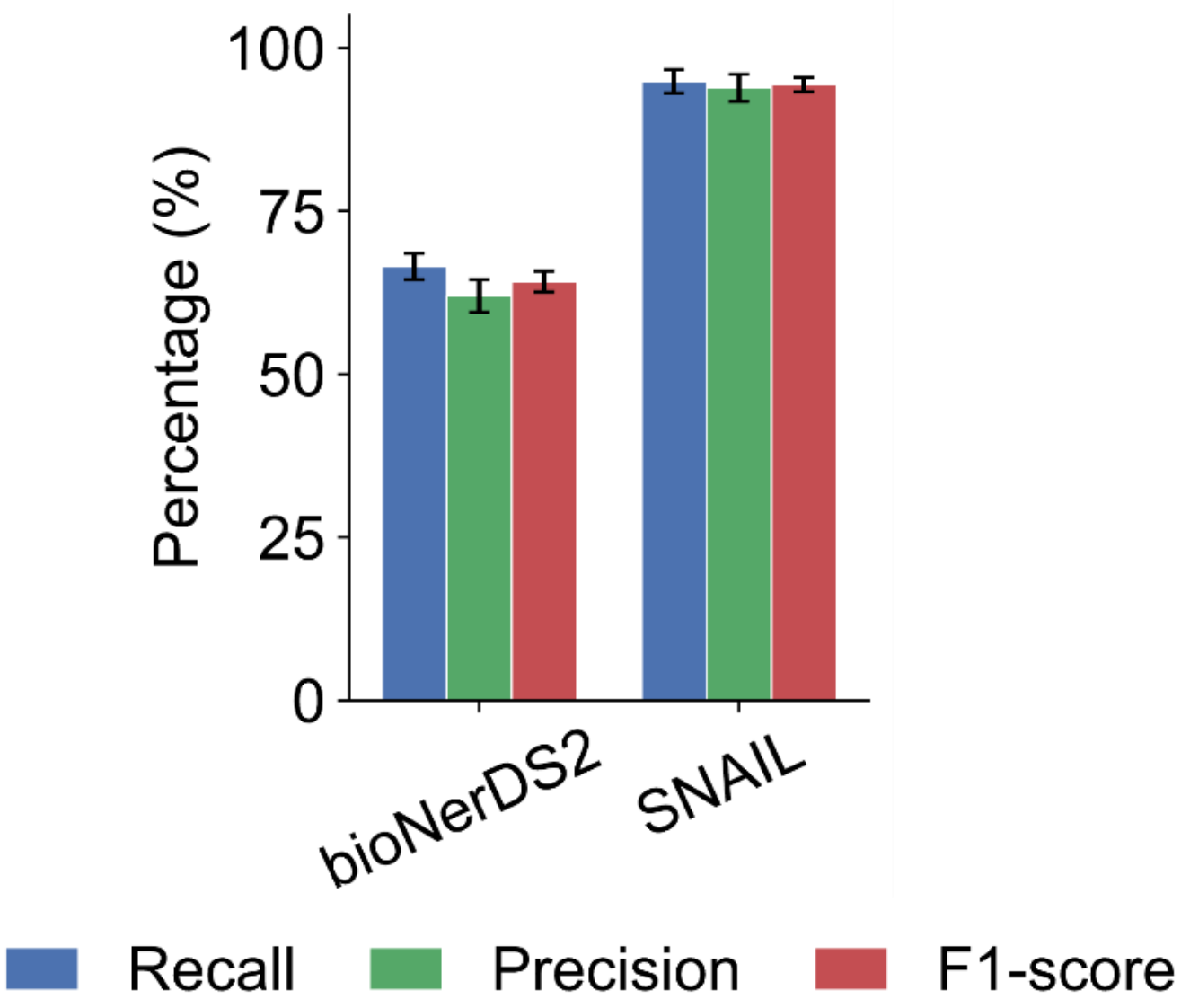

**Fig. S2: Benchmarking of SNAIL against bioNerDS2.** Performance comparison on benchmark datasets. Precision, recall, and F1-score of SNAIL and bioNerDS2 evaluated on DS2. Error bars indicate standard deviation across cross-validation folds.

**Table S1: Generational large language models utilized for benchmark comparisons.** Models are organized by institutional lineage to track the progression from earlier-generation variants to recent state-of-the-art updates. The selected models represent the core evolution of the Claude, Gemini, Grok, and GPT product ecosystems used to evaluate SNAIL performance.

| LLM Ecosystem | Earlier Generation | Latest Generation |
|---|---|---|
| **Claude** | Claude-3.5-Sonnet | Claude-Sonnet-4.6 |
| **Gemini** | Gemini-2.5-Flash | Gemini-3-Flash-preview |
| **Grok** | Grok-4 | Grok-4.20-0309-reasoning |
| **ChatGPT** | GPT-4.1-mini | GPT-5.4-mini |